\documentclass[letterpaper, 10 pt, conference]{ieeeconf}  

\IEEEoverridecommandlockouts                              

\usepackage{amsmath} 
\usepackage{amssymb}  
\usepackage{graphicx} 
\usepackage{float}
\usepackage{booktabs}
\usepackage{algorithm}
\usepackage{algorithmic}
\usepackage{xcolor}
\usepackage{makecell}
\usepackage{caption}
\usepackage{subcaption}
\usepackage{multirow}
\usepackage{cite}
\usepackage{url}
\title{\LARGE \bf
Gait-Adaptive Perceptive Humanoid Locomotion with Real-Time Under-Base Terrain Reconstruction
}

\author{
Haolin Song$^{1,2}$,
Hongbo Zhu$^{3,2}$,
Tao Yu$^{2}$, 
Yan Liu$^{4,2}$, 
Mingqi Yuan$^{5,2}$,
Wengang Zhou$^{1 \text{\textdagger}}$, 
Hua Chen$^{6,2 \text{\textdagger}}$,
Houqiang Li$^{1}$, 
\thanks{* Work done at LimX Dynamics. Project lead: Tao Yu}
\thanks{$\text{\textdagger}$ Corresponding author}
\thanks{$^1$ Department of Electronic Engineering and Information Science
(EEIS), University of Science and Technology of China, Hefei 230027,
China. Email: {\tt\footnotesize hlsong@mail.ustc.edu.cn,zhwg@ustc.edu.cn, lihq@ustc.edu.cn}}
\thanks{$^2$ LimX Dynamics, Shenzhen, China.}
\thanks{$^3$ Hong Kong University of Science and Technology, Hong Kong SAR, China. Email:  {\tt\footnotesize hzhubi@connect.ust.hk}}
\thanks{$^4$ School of Mechanics Engineering, Harbin Institute of Technology (HIT), Harbin Heilongjiang 150001, China. Email: {\tt\footnotesize liuyan98@stu.hit.edu.cn}}
\thanks{$^5$ Department of Computing, The Hong Kong Polytechnic University, Hong
Kong SAR, China Email: {\tt\footnotesize mingqi.yuan@connect.polyu.hk}}
\thanks{$^6$ Zhejiang University-University of Illinois Urbana-Champaign Institute (ZJUI), Haining, China. Email: {\tt\footnotesize huachen@intl.zju.edu.cn}}
}

\begin{document}

\maketitle
\thispagestyle{empty}
\pagestyle{empty}

\begin{abstract}
For full-size humanoid robots, even with recent advances in reinforcement learning-based control, achieving reliable locomotion on complex terrains, such as long staircases, remains challenging. In such settings, limited perception, ambiguous terrain cues, and insufficient adaptation of gait timing can cause even a single misplaced or mistimed step to result in rapid loss of balance. We introduce a perceptive locomotion framework that merges terrain sensing, gait regulation, and whole-body control into a single reinforcement learning policy. A downward-facing depth camera mounted under the base observes the support region around the feet, and a compact U-Net reconstructs a dense egocentric height map from each frame in real time, operating at the same frequency as the control loop. The perceptual height map, together with proprioceptive observations, is processed by a unified policy that produces joint commands and a global stepping-phase signal, allowing gait timing and whole-body posture to be adapted jointly to the commanded motion and local terrain geometry. We further adopt a single-stage successive teacher–student training scheme for efficient policy learning and knowledge transfer.
Experiments conducted on a 31-DoF, 1.65 m humanoid robot demonstrate robust locomotion in both simulation and real-world settings, including forward and backward stair ascent and descent, as well as crossing a 46 cm gap. Project Page \url{https://ga-phl.github.io/}
\end{abstract}

\section{INTRODUCTION}
Humanoid locomotion on complex terrains remains a central challenge for full-sized robots~\cite{tong2024advancements}. 
Compared with quadrupeds, humanoids must achieve precise foothold placement with a high center of mass and a small support polygon, which makes them highly sensitive to local terrain errors and poorly timed steps~\cite{xie2025humanoid}. 
Recent advances in reinforcement learning (RL) and simulation-to-real transfer have produced impressive bipedal and humanoid controllers that withstand large disturbances and track velocity commands on flat or mildly uneven terrain~\cite{legged-robots,li2025reinforcement,radosavovic2024real,gu2024humanoid}. 
However, when deployed on long staircases or gaps, such ``blind'' policies tend to behave more like robust fall-prevention controllers than deliberate planners of footholds and gait patterns.

These limitations highlight the need for exteroceptive perception and, equally importantly, for a tight integration of perception with gait timing and whole-body motion. 
For bipedal walking, the terrain directly beneath the base and around the feet is particularly critical, because its accurate estimation supports safe foothold selection and appropriate gait adaptation. 
The core question is therefore not only \emph{how} to sense terrain, but \emph{how} to encode and fuse that terrain information with gait and joint control in a way that remains trainable at scale and robust in deployment.

Existing perception-based locomotion pipelines still exhibit notable limitations. Forward-facing depth-camera policies~\cite{zhuang2024humanoid,luo2024pie,sun2025dpl} infer the unseen foot-sole terrain from short histories of images. Their narrow, forward-only field of view and reliance on temporal memory make them sensitive to noise and occlusions, causing them to lose track of the terrain under the base when the robot slows down, stops, or changes direction. 
LiDAR-based elevation-map methods~\cite{long2024learninghumanoid,wang2025beamdojo,yang2022real,hoeller2022neural} build robot-centric height maps by fusing LiDAR measurements with odometry, but the separate mapping and pose-estimation stack adds complexity and latency, and can suffer from drift and incomplete coverage near the feet. 

\begin{figure}[tb]
    \centering    \includegraphics[width=0.95\linewidth]{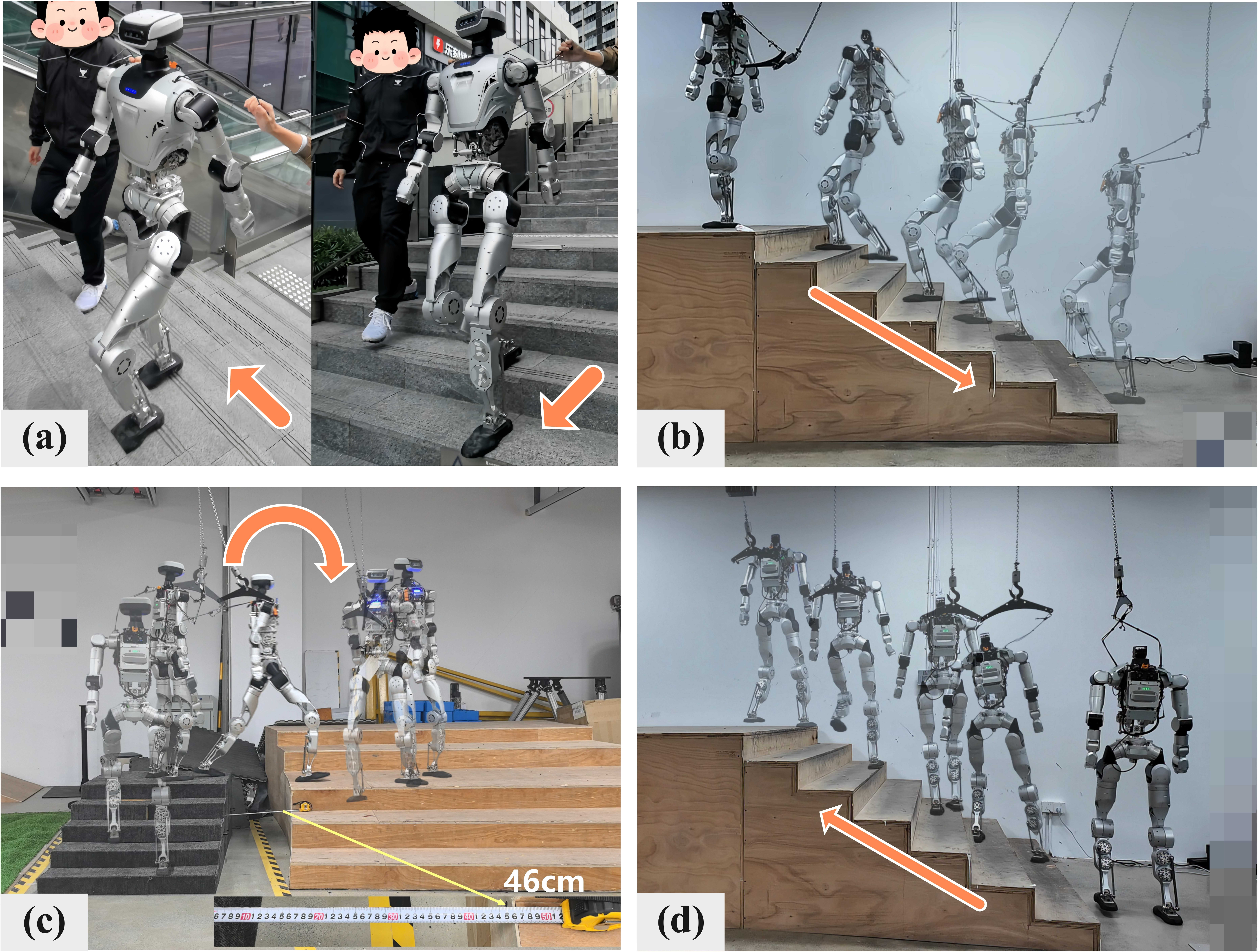}
    \setlength{\belowcaptionskip}{-15pt}  
    \caption{Full-sized humanoid robot Oli performing gait-adaptive locomotion on complex terrains: (a) climbing up and down long outdoor staircases; (b)  going down stairs backwards; (c) crossing a 46 cm gap; and (d)  climbing up stairs sideways}
    \label{fig:robot_real_1}
\end{figure}

Meanwhile, many locomotion controllers treat gait timing as an external signal. The step frequency is either pre-specified~\cite{gu2024advancing} or generated by an additional vision-based gait modulator~\cite{duan2024learning}. This decouples step frequency and phase from whole-body motion, weakening the coupling between terrain perception, gait timing, and joint commands, and limiting end-to-end optimization of gait to the current state and local terrain.

In this work, we aim to close these gaps by designing a perceptive locomotion framework with adaptive gait control.
We first introduce an under-base terrain perception module specifically designed for humanoid walking. A downward-looking depth camera mounted under the base captures the terrain beneath the base and around the feet, which is the key region for safe stepping in omnidirectional motion. Because these images are strongly affected by self-occlusions from the body and legs, we employ a lightweight U-net that converts each single depth frame into a dense, egocentric height map. This approach avoids multi-sensor fusion and explicit temporal mapping, while still providing a controller-friendly local terrain representation at the control frequency.

On top of this perception, we propose a unified policy that simultaneously outputs whole-body joint targets and a scalar gait-frequency action. Instead of following an externally prescribed gait schedule, the policy learns to regulate step timing jointly with body motion from the same proprioceptive and perceptual inputs. This end-to-end form enables the controller to adjust its gait rhythm automatically in response to commanded motion and local terrain, resulting in more deliberate stepping on stairs and gaps, as well as faster and smoother walking on flat ground.

Finally, we employ Successive Teacher–Student (S-TS), a single-stage training scheme to transfer knowledge from privileged to partial observations in a stable and data-efficient manner. 
A privileged teacher first learns a strong locomotion policy, while a student encoder is supervised to match the teacher’s latent features from noisy, partial observations.
A switch gate then gradually transfers environment interaction from the teacher to the student, so that the final deployable policy is trained under realistic partial inputs while still being guided by the teacher’s representation.

Our main contributions can be summarized as follows:
\begin{itemize}
\item A terrain perception module that reconstructs a local under-base height map at 50\,Hz from a single downward-looking depth frame.
\item A unified policy that jointly outputs whole-body joint targets and gait frequency, enabling tightly coupled, terrain-aware, gait-adaptive humanoid locomotion.
\item A single-stage successive teacher–student framework for efficient knowledge learning and transfer from privileged to partial observations.
\item Comprehensive validation on a full-sized humanoid robot, demonstrating omnidirectional walking and terrain-aware gait adaptation on diverse simulated and real stair and gap terrains.
\end{itemize}

\begin{figure*}[htb]
    \centering
    \includegraphics[width=0.99\textwidth]{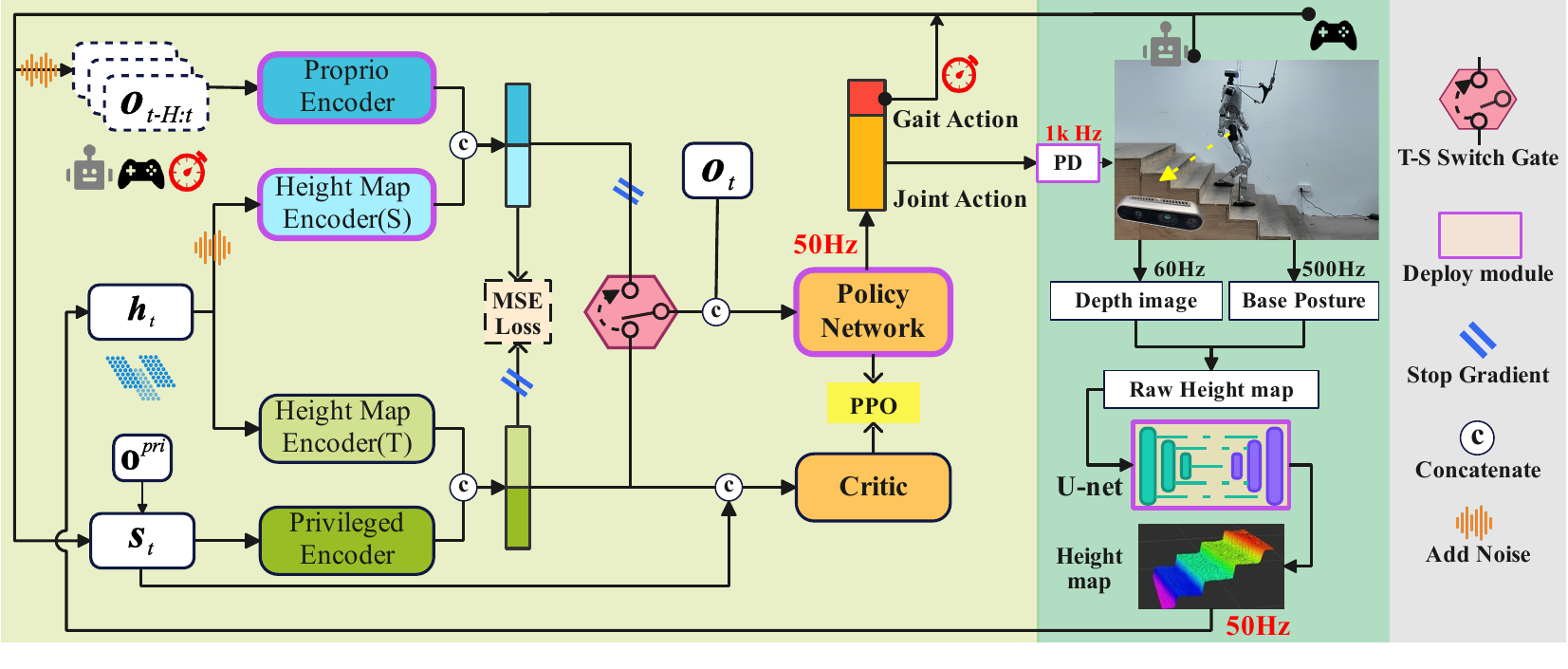}
    \setlength{\belowcaptionskip}{-10pt}  
    \caption{Overview of the proposed Successive Teacher–Student (S-TS) framework and deployment pipeline. A teacher–student switch gate gradually transfers rollouts from the privileged teacher to the student. The unified policy outputs both joint actions and a scalar gait-frequency action. A downward-looking depth image is converted into an under-base height map by the perception module, which runs at 50 Hz together with the control policy.}
    \label{fig:pipeline}
\end{figure*}

\section{Related Work}
\subsection{Legged Locomotion Control}
A large body of work on humanoid and biped locomotion studies controllers that rely solely on proprioceptive feedback~\cite{radosavovic2024real,long2024learninghumanoid,siekmann2021blind}. 
These policies are typically trained to follow velocity commands and withstand disturbances on relatively simple terrain; they struggle to reason about foot placement and step timing on long staircases or wide gaps.
To regularize motion, many approaches introduce an explicit gait phase or frequency that drives predefined periodic patterns~\cite{gu2024advancing,xue2025unified,wang2025more}. 
The step rate is often either fixed or computed as a simple function of the commanded speed. This yields stable periodic motions on flat terrain, but makes it hard to change the gait rapidly according to different terrain.
Recent learning-based methods embed a gait-phase action into the same policy as joint targets~\cite{lu2025contrastive}; however, modulation remains purely proprioception-driven and is primarily used to avoid falls under disturbances. Consequently, the gait signal is still not optimized jointly with terrain perception.

\subsection{Perceptive Humanoid Locomotion}
Perceptive locomotion enhances leg control with exteroceptive sensing, typically using cameras or LiDAR. 
Camera-based approaches process short RGB-D or depth snippets from forward-facing sensors to infer nearby traversable terrain and footholds~\cite{zhuang2024humanoid,luo2024pie,sun2025dpl}, but the limited viewing direction and reliance on memory make the terrain in the stance region quickly outdated when the robot turns, slows, or pauses, which hampers sideways and backward motion on stairs and gaps.
LiDAR-based elevation-map methods maintain robot-centered height maps by integrating range scans with odometry and related estimates~\cite{long2024learninghumanoid,wang2025beamdojo}. 
They offer broader coverage but require a separate mapping pipeline with motion estimation, introducing extra complexity, delay, and occlusions close to the torso and feet, where accurate geometry is most critical.

Several perceptive frameworks additionally attach a vision-driven module that adjusts step rate or phase, while the main controller focuses on command tracking ~\cite{duan2024learning}, so contact timing is only weakly tied to the perceptual stream. 
In contrast, we use a downward-looking depth sensor and a lightweight reconstruction network to obtain a dense under-base height map from a single frame, and feed this local representation directly into a unified policy that outputs both joint actions and a gait-frequency signal.

\section{Method}
An overview of the Successive Teacher–Student (S-TS) architecture for adaptive humanoid perceptive locomotion is shown in Fig. \ref{fig:pipeline}. 
We employ a single-stage training framework to train the teacher and student via an asymmetric Actor-Critic approach. The teacher and student share the same policy head $\pi_{\theta}$ and critic network $V_{\phi}$, with the only difference being the observation encoder.
\subsection{Observation Space}
We consider three types of observations in our perceptive humanoid locomotion framework: proprioceptive (\(o^{\text{pro}}\)), privileged (\(o^{\text{pri}}\)), and exteroceptive (\(o^{\text{per}}\)). The teacher and critic use noise-free observations from all three types, while the student relies only on noisy proprioceptive and perceptual inputs, obtained by injecting Gaussian noise into the teacher's observations. The privileged encoder takes the state $s_t^{\text{pri}} = [o_t^{\text{pro}}, o_t^{\text{pri}}]$ as input, while the proprio encoder uses a historical sequence of proprioceptive observations $o_t^{\text{his}} = [o_{t-H+1}^{\text{pro}}, \dots, o_t^{\text{pro}}]$.

The proprioceptive observation $o_t^{\text{pro}}$ comprises user commands $\boldsymbol{c}_t=[v_x, v_y, \omega_{\text{yaw}}]$, the body angular velocity $\boldsymbol{\omega}_t$, projected gravity $\boldsymbol{g}_t$, joint positions $\boldsymbol{q}_t$, joint velocities $\dot{\boldsymbol{q}}_t$, previous actions $\boldsymbol{a}_{t-1}$,  and the gait signals \(\{f_t, \sin(2\pi\phi_t), \cos(2\pi\phi_t)\}\).

The privileged observation $o_t^{\text{pri}}$ contains additional information, including base linear velocity $\boldsymbol{v}_t$, joint torques $\boldsymbol{\tau}_t$, joint accelerations $\ddot{\boldsymbol{q}}_t$, foot contact forces $\boldsymbol{F}_t$, foot heights $\boldsymbol{h}_{f,t}$, and base height $h_{b,t}$.

The perceptual observation $o_t^{\text{per}}$ is a local egocentric height map $\boldsymbol{h}_t \in \mathbb{R}^{425}$ describing the terrain beneath and around the robot's feet.

\subsection{Unified Action Space}

We propose a unified policy in which a single policy network simultaneously outputs joint commands and gait parameters, enabling real-time coordination of motion control and gait modulation within a single, coherent policy. Concretely, the network produces a 32-dimensional action vector
$a_t = [a_t^{\text{joints}}, f_t]$,
where $a_t^{\text{joints}} \in \mathbb{R}^{31}$ specifies target positions for all joints, and $f_t$ is a scalar gait frequency that controls a global gait phase $\phi_t$. The phase is updated as
$$\phi_t = \mathrm{mod}(\phi_{t-\Delta t} + \Delta t \cdot f_t, 1.0)$$
with $\Delta t$ denoting the control timestep. 
Intuitively, a larger $f_t$ accelerates the stepping cycle, while a smaller $f_t$ slows it down. The left and right legs maintain a fixed phase offset of $0.5$ to ensure standard alternating steps.

To improve stability and smoothness, the raw gait frequency output is post-processed before updating $\phi_t$. Specifically, $f_t$ is first scaled and clipped into a feasible range to prevent excessively slow or fast stepping, and then passed through a short-term averaging filter that suppresses abrupt changes between consecutive control steps. This filtering stabilizes the evolution of the gait phase while still allowing the policy to adapt the stepping rate over time.

By embedding $f_t$ into the same action vector as joint targets, the policy can reason jointly about \emph{when} to step and \emph{how} to configure the body over the gait cycle. This unified, end-to-end formulation eliminates the need for a separate gait generator, allowing reinforcement learning to directly shape both timing and joint motions, thereby improving adaptability across diverse terrains and commanded speeds.

\subsection{Successive Teacher-Student Architectures}
We propose a Successive Teacher-Student (S-TS) framework in which a Teacher-Student Switch Gate controls the relative participation of teacher and student throughout training. In the early stage, a teacher-exclusive interaction mode is used: only the teacher interacts with the environment and generates trajectories. The teacher encoder and shared policy network are jointly updated using privileged information and a stable optimization target, leading to fast and stable convergence.
During this stage, the student does not affect the environment and instead learns via a supervised auxiliary task, mapping its own noisy and partial observations to the teacher's latent representations to compensate for the absence of privileged information. As training progresses, the Switch Gate gradually increases the student's share of environments, leading to a parallel mode of teacher–student interaction. The shared policy is optimized on trajectories from both agents, preserving the teacher's well-shaped behavior while adapting to the student's noisy deployment observations, and ultimately yielding a robust, environment-adaptive locomotion policy.
\begin{algorithm}[tb]
\caption{Successive Teacher-Student (S-TS) Training}
\label{alg:pipeline}
\begin{algorithmic}[1]
\STATE Initialize teacher encoder $E^T(\theta_E^T)$, student encoder $E^S(\theta_E^S)$, shared policy $\pi(\theta_\pi)$, and value network $V(\phi)$
\STATE Initialize $N$ parallel environments (all assigned to the teacher at iteration $k=0$)

\FOR{$k = 0,1,\dots$} 
    \STATE Update student ratio $\lambda_k \in [0,1]$, assign $(1-\lambda_k)N$ teachers and $\lambda_k N$ students.
    \STATE Collect teacher $\mathcal{D}^T$ and student $\mathcal{D}^S$

    \STATE Update policy and value networks:
    $$
    L^{\text{ppo}} = L^{\text{ppo-T}}(\theta_E^T, \theta_\pi \mid \mathcal{D}^T)
                   + L^{\text{ppo-S}}(\theta_\pi \mid \mathcal{D}^S)
    $$
    $$
    \theta_\pi \leftarrow \theta_\pi + \alpha_{\text{ppo}} \nabla_{\theta_\pi} L^{\text{ppo}}
    $$
    $$
    \phi \leftarrow \phi - \alpha_{\text{ppo}} \nabla_{\phi} L^{\text{value}}
    $$

    \STATE Update student encoder via reconstruction loss:
    $$
    \theta_E^S \leftarrow \theta_E^S - \alpha_{\text{ts}}
    \nabla_{\theta_E^S} L^{\text{rec}}(\theta_E^S)
    $$

    \STATE \textit{// When $\lambda_k = 0$, $\mathcal{D}^S$ is empty and the update reduces to teacher-only PPO.}
\ENDFOR
\end{algorithmic}
\end{algorithm}

\subsection{Network Architecture and Loss}

The teacher encoder comprises a privileged encoder $E^{\text{pri}}$ and a perception encoder $E^{\text{perT}}$ that process the privileged state $s^{\text{pri}}$ and the complete height map $h^{\text{T}}$:
$z^{\text{T}} = (E^{\text{pri}}(s^{\text{pri}}), E^{\text{perT}}(h^{\text{T}}))$.
The student encoder comprises a proprioceptive encoder $E^{\text{pro}}$ and a perception encoder $E^{\text{perS}}$ that process the historical proprioceptive observations $o^{\text{his}}$ and the noisy height map $h^{\text{S}}$:
$z^{\text{S}} = (E^{\text{pro}}(o^{\text{his}}), E^{\text{perS}}(h^{\text{S}}))$.
For both agents, actions are generated as $a_t = \pi_{\theta}(z_t, o_t^{\text{pro}})$.
All networks are implemented as MLPs with ELU activations.
We employ Proximal Policy Optimization (PPO)~\cite{schulman2017proximal} for policy learning. During training, the Teacher-Student Switch Gate controls the student ratio $\lambda \in [0,1]$: when $\lambda = 0$, all environments are teacher-controlled (teacher-exclusive phase); when $\lambda > 0$, teacher- and student-controlled environments run in parallel. We separately record teacher trajectories $\mathcal{D}^T$ and student trajectories $\mathcal{D}^S$, and define the corresponding PPO losses as:
$$
L^{\text{ppo-T}}(\theta_E^{\text{T}}, \theta_{\pi}) =
\mathbb{E}_t \big[
    \min(r_t A_t,\,
        \text{clip}(r_t, 1-\epsilon, 1+\epsilon) A_t)
    \,\big|\, \mathcal{D}^T
\big],
$$
$$
L^{\text{ppo-S}}(\theta_{\pi}) =
\mathbb{E}_t \big[
    \min(r_t A_t,\,
        \text{clip}(r_t, 1-\epsilon, 1+\epsilon) A_t)
    \,\big|\, \mathcal{D}^S
\big],
$$
where $r_t$ is the probability ratio and $A_t$ the advantage (estimated with GAE). In the teacher-exclusive phase ($\lambda = 0$), the policy is updated only with $L^{\text{ppo-T}}$. In the parallel phase ($\lambda > 0$), we optimize the combined objective
$$
L^{\text{ppo}}(\theta_{\pi}) =
L^{\text{ppo-T}}(\theta_E^{\text{T}}, \theta_{\pi} \mid \mathcal{D}^T)
+
L^{\text{ppo-S}}(\theta_{\pi} \mid \mathcal{D}^S).
$$
To bridge the gap between privileged and noisy observations, we align the student's latent representation with the teacher's using a mean squared error loss.
$$
L^{\text{rec}}(\theta_E^{\text{S}}) =
\mathbb{E}_t [\| z^{\text{T}} - z^{\text{S}} \|_2^2 ].
$$
In addition, a mirror loss $L^{\text{mir}}$ enforces action symmetry under mirrored robot and terrain states. Training details are summarized in Algorithm~\ref{alg:pipeline}.

\subsection{Single-frame Height Map Reconstruction}
We employ a perception module that focuses on the terrain directly beneath the base, which is the most relevant region for foothold selection and omnidirectional walking. A downward-looking depth camera is mounted under the base. However, the resulting depth images are strongly affected by self-occlusions from the legs, so they cannot be fed to the locomotion policy in their raw form.

\begin{figure}[b]
    \centering
    \includegraphics[width=0.95\linewidth]{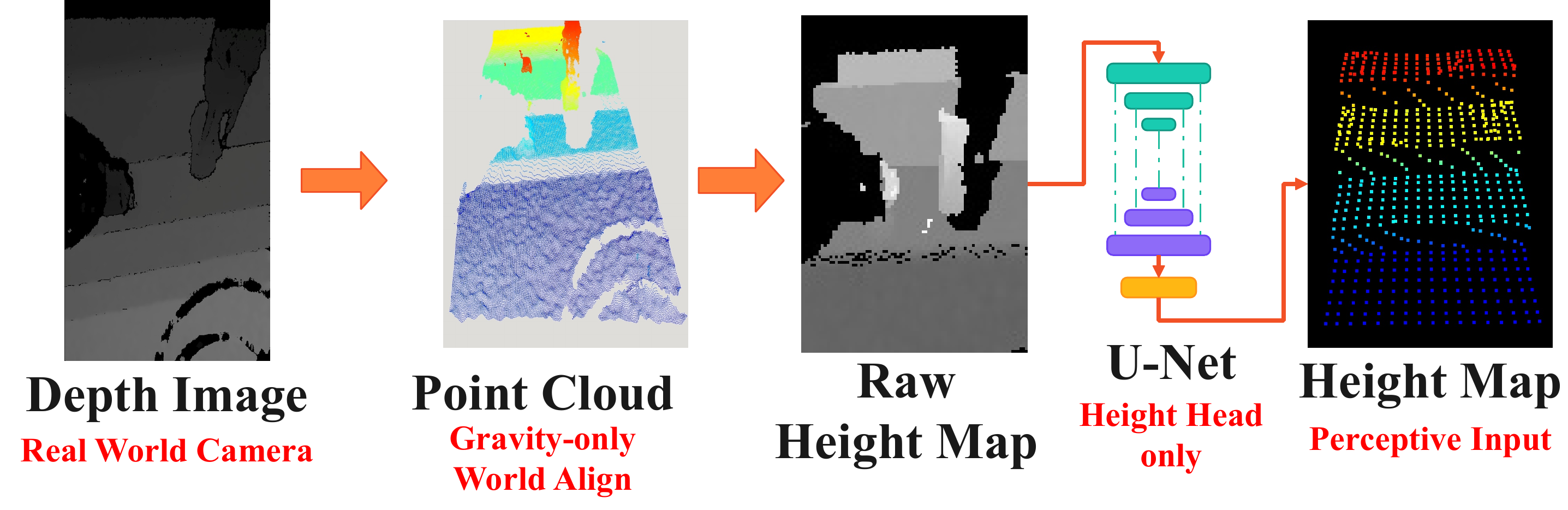}
    \caption{This figure illustrates the pipeline of single-frame height map reconstruction using a U-Net model in deployment.}
    \label{fig:real-camera}
\end{figure}

Fig.~\ref{fig:real-camera} illustrates the under-base perception module. The input is a single depth image $I_t$ from the downward-facing camera. We first transform $I_t$ into a gravity-aligned point cloud using the camera intrinsics and extrinsics. Points within a fixed window around the robot are then selected and projected onto the horizontal plane to form a raw height map $\hat{H}^{\text{raw}}_t$. This map already encodes local terrain, but contains holes and artifacts that are self-occluded by the robot.
\begin{figure}[tb]
    \centering
    \setlength{\belowcaptionskip}{-15pt} \includegraphics[width=0.95\linewidth]{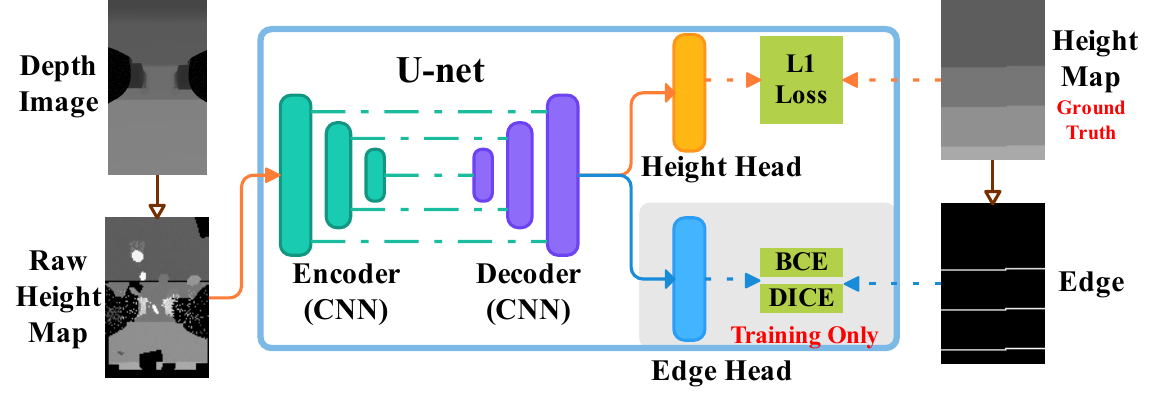}
    \caption{U-Net-based single-frame heightmap reconstruction network. 
The depth image is converted to a noisy base-centric heightmap and processed by a U-Net with two heads: 
A height head supervised by L1 loss and an edge head (training only) using BCE and Dice losses.}
\label{fig:U-net}
\end{figure}

To obtain a usable representation, $\hat{H}^{\text{raw}}_t$ is processed by a U-Net encoder and decoder. From the shared latent features, two output branches are attached: a \emph{height head} that predicts a refined height map $\hat{H}^{\text{height}}$, and an \emph{edge head} that produces an edge map $\hat{E}^{\text{edge}}$ highlighting height discontinuities such as stair fronts or curb edges. The edge branch is used as an auxiliary task, guiding the network to recover and preserve sharp terrain boundaries, especially near regions that are partially missing due to self-occlusion, which would otherwise be over-smoothed by pure height regression.

As shown in Fig. \ref{fig:U-net}, training is formulated as a multi-task learning problem. For height prediction, we use an $L_1$ loss.
$$
L_{\text{height}} = \|\hat{H}^{\text{height}} - H^{\text{truth}}\|_1,
$$
where $H^{\text{truth}}$ denotes the ground-truth local height map. For the edge branch, we construct a binary edge target $E^{\text{truth}}$ by applying an edge detector to $H^{\text{truth}}$, and supervise it with the sum of Binary Cross-Entropy and Dice losses~\cite{li2020dice}: 
$$
L_{\text{edge}} =
L_{\text{BCE}}(\hat{E}^{\text{edge}}, E^{\text{truth}})
+ L_{\text{Dice}}(\hat{E}^{\text{edge}}, E^{\text{truth}}).
$$
The overall training objective is
$$
L_{\text{total}} = L_{\text{height}} + \lambda_{\text{edge}} L_{\text{edge}},
$$
with $\lambda_{\text{edge}}$ controlling the strength of the auxiliary edge supervision.

At run time, only $\hat{H}^{\text{height}}$ is forwarded to the locomotion policy as the perceptive input, while the edge head is discarded. Because the network operates on a single depth frame without temporal aggregation or additional sensors, it remains lightweight and can be executed at the control frequency, providing an up-to-date under-base height map in all directions around the feet.

\subsection{Reward Design}

We primarily adopt the locomotion reward from the IsaacLab official implementation. For gait control, we utilize the rewards from ~\cite{xue2025unified} to regulate both upper- and lower-body motion, resulting in a humanoid-like walking pattern. Inspired by ~\cite{zhuang2024humanoid, wang2025beamdojo}, we further add foot-placement rewards to encourage safer footholds on complex terrains. Our policy network outputs both joint actions and a gait frequency, enabling adaptive gait control. For the gait frequency, we reuse the original regularization on joint actions: an action-smoothness term that encourages gradual changes, together with an action-limit term that constrains the frequency to the range $[0.7, 1.3]$ to maintain a reasonable frequency. Together, these rewards allow the robot to maintain a humanoid gait and precise foot placement across a wide variety of challenging environments. A detailed summary of all reward terms is provided in Table~\ref{tab:reward}.

\section{EXPERIMENTS}
\subsection{Robot Platform}

\begin{figure}[t]
    \centering
    \includegraphics[width=0.6\columnwidth]{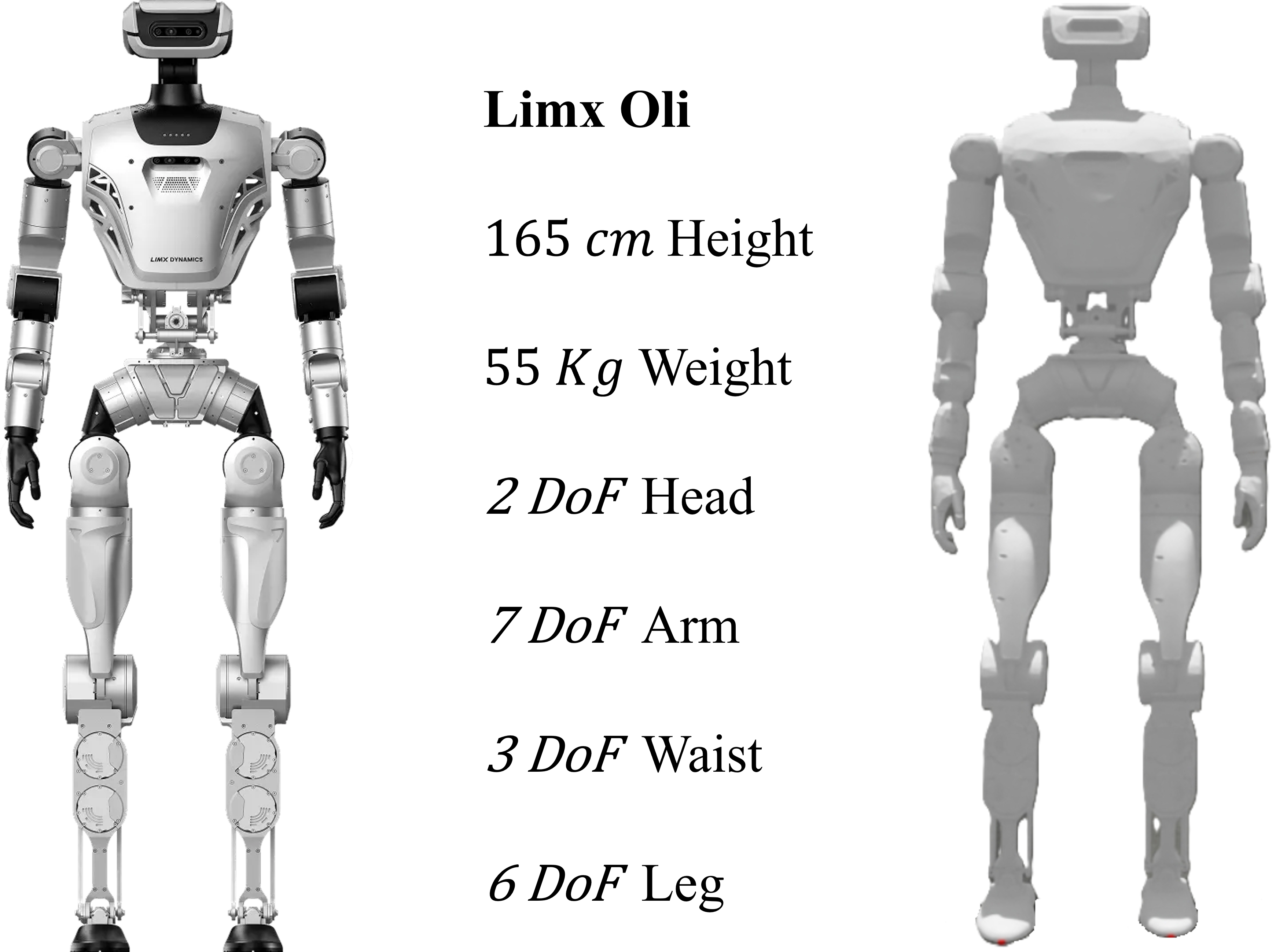}
    \setlength{\belowcaptionskip}{-15pt} 
    \caption{Robot hardware ``Limx Oli'' in real-world (left) and simulation (right) setups, illustrating the robot's physical dimensions and degrees of freedom.}
    \label{fig:robot}
\end{figure}

We use a full-sized humanoid robot, ``Limx Oli'' ~\cite{oli}, for all simulation and real-world experiments. As shown in Fig.~\ref{fig:robot}. The robot weighs 55 kg, stands 1.65\,m tall, and has 31 actuated degrees of freedom (DoFs): 6 in each leg, 7 in each arm, 3 in the waist, and 2 in the head. We actively control all 31 DoFs, which significantly increases the control complexity and coordination difficulty. 
For onboard computation, the robot is equipped with an NVIDIA Jetson Orin NX, and for perception, it uses an Intel RealSense D435i RGB-D camera. The depth images from the camera are used to reconstruct local height maps, which are then fed into the locomotion policy, 
as shown in Fig. \ref{fig:real-camera}

\begin{table}[htb]
\centering
\caption{Reward Term}
\label{tab:reward}
\setlength{\tabcolsep}{3pt}
\renewcommand{\arraystretch}{1.1}
\begin{tabular}{lcc}
\toprule
Reward Term & Formula  & weight \\
\midrule
Lin. velocity track & $\exp\!\big(-4\|{\bf v}_{xy}^{\rm cmd}-{\bf v}_{xy}\|^2\big)$ & 1.0 \\
Ang. velocity track & $\exp\!\big(-4(\omega_{z}^{\rm cmd}-\omega_{z})^2\big)$ & 0.5 \\
\makecell[l]{Base height \\ (w.r.t.\ feet) }
& $\exp\!\big(-200(h^{\rm tgt}-h)^2\big)$ & 0.4 \\
\midrule
Contact--swing track & \makecell[c]{$-\sum_i\bar C_i\!\left[1-\exp(\|f^{\text{foot}}_i\|^2/50)\right]$\\
$-\sum_i C_i\!\left[1-\exp(\|v^{\text{foot}}_{xy,i}\|^2/5)\right]$} & 0.5 \\
Natural swing arm & \makecell[c]{$\exp\!\big((q_{\rm arm}-q^{\rm tgt}_{\rm arm})^2/0.02\big)$\\
$+\;\exp\!\big((v_{\rm arm}-v^{\rm tgt}_{\rm arm})^2/0.1\big)$} & 0.05 \\
\midrule
Action smoothness & $\|{\bf a}_t-2{\bf a}_{t-1}+{\bf a}_{t-2}\|^2$ & $-2.5$e-03 \\
Gait action limit & {-$n_{\rm lim}$} & -0.25 \\
Joint accel.\ L2 & $\|\ddot{\bf q}\|^2$ & $-5$e-07 \\
Joint vel.\ L2 & $\|\dot{\bf q}\|^2$ & $-1$e-03 \\
Joint torque L2 & $\|{\boldsymbol\tau}\|^2$ & $-4$e-07 \\
Torque rate & $\|{\boldsymbol\tau}_t-{\boldsymbol\tau}_{t-1}\|^2$ & $-1.5$e-07 \\
Joint power & $|\boldsymbol\tau|^\top|\dot{\bf q}|$ & $-2.5$e-07 \\
\makecell[l]{Joint limits \\ (pos/vel/tor) }
& \makecell[c]{-$n_{\rm lim}$} & 0.2/0.025/0.01 \\
\makecell[l]{Joint deviation \\(waist/arm/hip)}
 & \makecell[c]{$-\sum|\theta_i-\theta^{\rm def}_i|^2$} & 0.3/0.01/0.5 \\
\midrule
Lin.\ accel.\ L2 & $\|\ddot{\bf v}\|^2$ & $-2$e-03 \\
Ang.\ vel.\ $xy$ L2 & $\|\boldsymbol\omega_{xy}\|^2$ & $-0.15$ \\
Proj.\ gravity L2 & $\|{\bf g}_x\|^2+\|{\bf g}_y\|^2$ & $-0.15$ \\
\midrule
Undesired contacts & $n_{\rm coll}$ & $-1.5$ \\
Feet stumble & $\mathbb{I}[\|{\bf F}_{\rm hor}\|>2\|{\bf F}_{\rm vert}\|]$ & $-1.5$ \\
Feet slide & $\mathbb{I}_{\rm c}(\|{\bf v}_{\rm foot}\|+0.25\|\boldsymbol\omega_{\rm foot}\|)$ & $-0.05$ \\
Feet air time & $\min({\rm air\_time},0.5)$ & 0.03 \\
Feet hold & $\exp\!\big(-100\|{\bf p}_{\rm foot}-{\bf p}_{\rm ctr}\|^2\big)$ & 0.5 \\
Feet stair flat & $\exp(-4\,r_D)$ & 0.25 \\
\bottomrule
\end{tabular}
\vspace{-2mm}
\end{table}

\subsection{Training and Deployment}
\subsubsection{Control Policy}
We conduct all policy training in the IsaacLab~\cite{mittal2025isaac}, which enables massively parallel reinforcement learning for robotics. Our locomotion environment runs 4096 parallel humanoid instances. We follow the curriculum setup provided by IsaacLab, where the robot is exposed to a variety of terrains, including flat ground, rough terrain, stepping stones, gaps, and pyramid stairs (up and down). At the beginning of each episode, the robot's pose is randomized near the center of the terrain. Every 10\,s, we resample the commanded body velocity, with
$\dot{x} \in [-1.0, 1.0]$\,m/s, 
$\dot{y} \in [-0.3, 0.3]$\,m/s, and 
$\dot{\psi} \in [-1.0, 1.0]$\,rad/s, 
encouraging forward/backward walking, lateral motion, and turning for omnidirectional locomotion.
To narrow the sim-to-real gap, we perform extensive domain randomization during training. We randomize the mass of the base, waist, and legs; the center of mass and inertia of the base; contact friction and restitution; PD gains; and external perturbations applied as random impulses and forces. This improves the robustness of the learned controller when deployed on the real robot. At deployment, the control policy runs at 50\,Hz, and its joint commands are tracked by PD controllers operating at 1\,kHz.

\subsubsection{Reconstruction Model Training and Deployment}

To collect training data for the height-map reconstruction module, we roll out a trained locomotion policy in IsaacLab with 100 parallel environments. A virtual depth camera is rigidly mounted on the underside of the floating base, pointing vertically downward to match the real hardware setup. The camera covers an effective field of view of $2.0\,\text{m} \times 1.0\,\text{m}$ on the ground, within which we define the local height map used by the policy as a robot-centric patch of $1.2\,\text{m} \times 0.8\,\text{m}$ around the base, discretized into a fixed grid with $5\,\text{cm}$ spatial resolution.

As the robot walks over diverse terrains, we record the base posture, the depth image, and the corresponding ground-truth local height map at $10\,\text{Hz}$, yielding a dataset of 10{,}000 frames. Each depth image is converted into a raw height map using the procedure described above, and additional noise is injected to mimic sensor artifacts and calibration errors, improving the robustness of the reconstruction model. The processed dataset is then used to train the U-Net-based reconstruction network.

In real-world deployment (Fig.~\ref{fig:real-camera}), the onboard downward-facing depth camera streams at $60\,\text{Hz}$. The reconstruction module processes each frame in about $11\,\text{ms}$ and provides height maps to the locomotion controller at $50\,\text{Hz}$, matching the control frequency.
\begin{figure}[tb]
    \centering    \includegraphics[width=0.8\columnwidth]
    {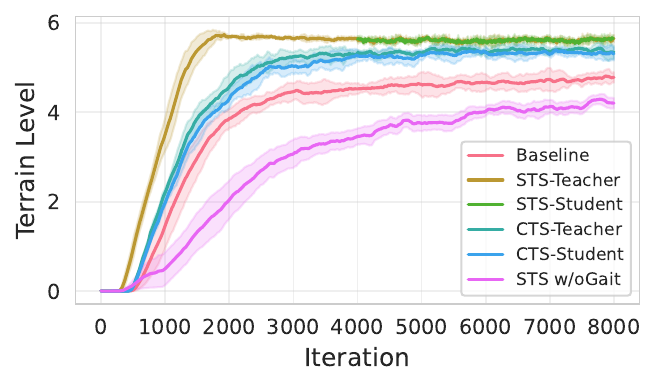}
    \setlength{\belowcaptionskip}{-15pt} 
    \caption{Training progress of different algorithms, showing the average terrain level achieved during training, with shaded areas representing standard deviation.}
    \label{fig:terrain_level}
\end{figure}

\subsection{Training Framework Results}
To evaluate the proposed Successive Teacher–Student framework, we compare:

\begin{itemize}
    \item \textbf{Baseline}: student policy trained with PPO only.
    \item \textbf{STS}: Our Successive Teacher–Student framework.
    \item \textbf{CTS}:~\cite{wang2024cts} concurrent teacher–student training, where teacher and student jointly update the shared policy from the beginning.
    \item \textbf{STS w/o Gait}: STS without the gait clock.
\end{itemize}

All methods use the same network architecture and hyperparameters and are trained for $8{,}000$ iterations. In STS, all $4{,}096$ environments are assigned to the teacher at the start. After $4{,}000$ iterations, we gradually increase the student ratio $\lambda$ from $0$ to $0.5$, so that at most half of the environments are controlled by students while the remaining half continue to follow the teacher, allowing students to gain experience under a strong teacher policy.
\begin{figure}[tb]
    \centering
    \includegraphics[width=0.95\linewidth]{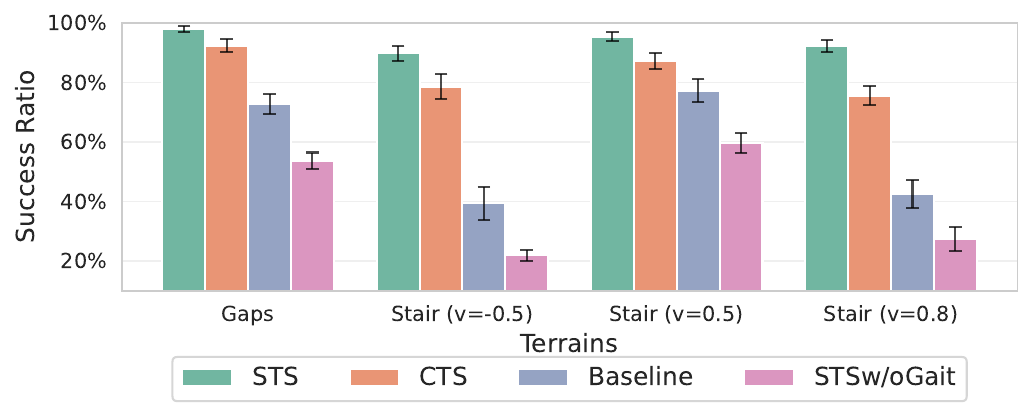}
    \setlength{\belowcaptionskip}{-15pt} 
    \caption{Success ratio comparison for different algorithms across various terrain types, including stairs and gaps, with varying speeds.}
    \label{fig:success_ratio}
\end{figure}

\begin{figure}[hb]
    \centering    \includegraphics[width=0.95\linewidth]{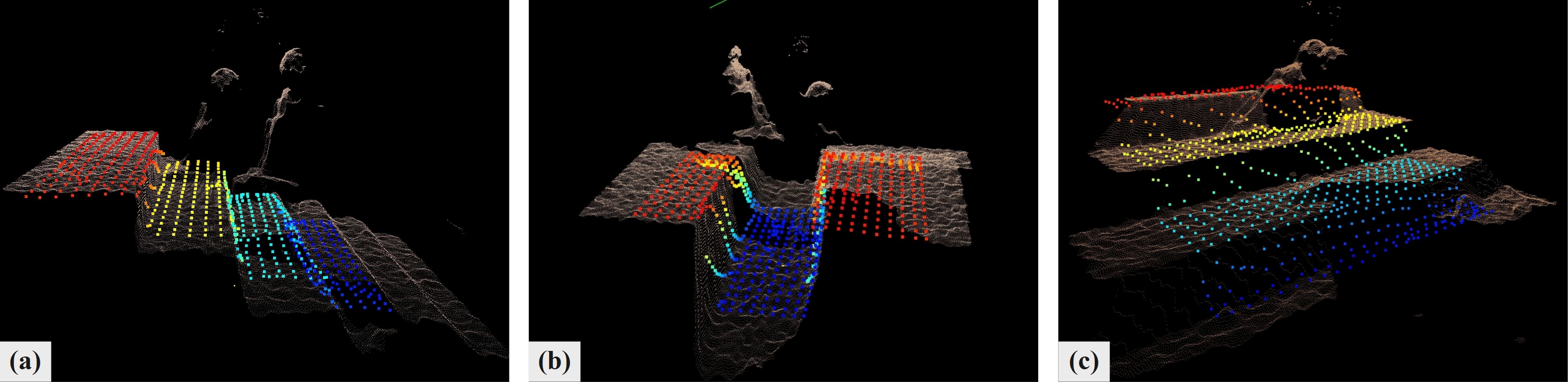}
    \caption{reconstruction results of the perception module on (a) frontal stairs, (b) gap terrain. (c) side-facing stairs,
    }
    \label{fig:pcd}
\end{figure}

\begin{figure*}[htb]
    \centering
    \includegraphics[width=0.9\textwidth]{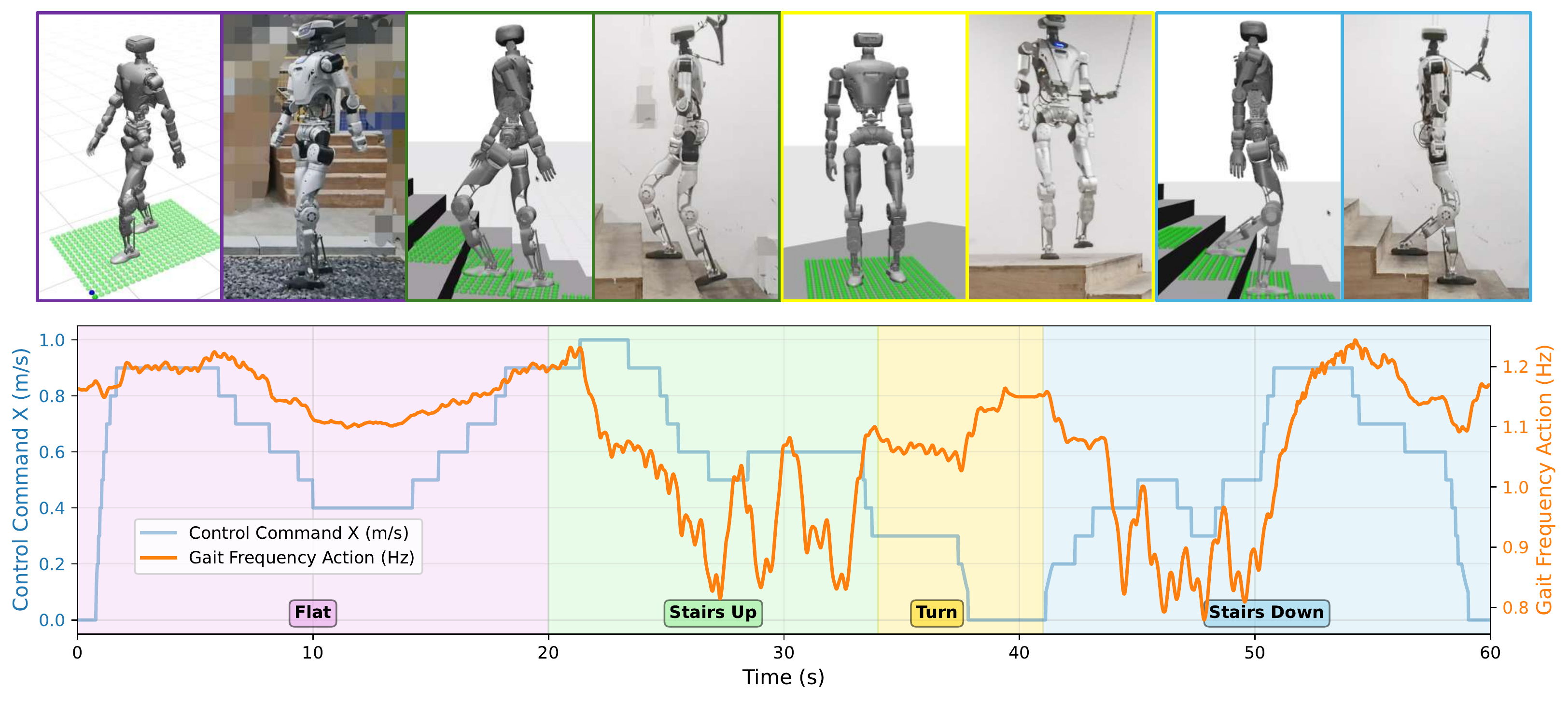 }
    \setlength{\belowcaptionskip}{-15pt}  
    \caption{Adaptive gait behavior in simulation (MuJoCo) and on the real robot across different terrains and walking speeds. The blue curve shows the forward command (yaw-rate commands are omitted for clarity), and the orange curve shows the post-processed gait frequency received by the robot.}
    \label{fig:gait_adaptation}
\end{figure*}

Fig.~\ref{fig:terrain_level} shows the evolution of the terrain level during training, which serves as a proxy for overall skill (higher levels correspond to more difficult terrains). STS-Teacher rapidly climbs to high curriculum levels by exploiting privileged information. Once students are introduced, STS-Student continues improving and eventually approaches the teacher's terrain level. CTS, in contrast, learns more slowly and converges to a lower final level, indicating that simultaneous teacher–student updates in the early phase introduce conflicting gradients. The Baseline and STS w/o Gait curves saturate at even lower terrain levels, showing the benefit of both privileged guidance and explicit gait control.

To further evaluate robustness across terrains in simulation, we measure the \emph{success ratio} on several representative scenarios (Fig.~\ref{fig:success_ratio}). For each terrain and commanded forward velocity, the success ratio is defined as the fraction of rollouts in which the robot travels a fixed distance without falling or violating safety constraints. We test a 15cm stair terrain at $v = -0.5$\,m/s, a 20cm stair terrain at $v \in \{0.5, 0.8\}$\,m/s, and a 40cm gap terrain, covering backward motion, moderate speeds, and highly challenging obstacles. Across all these settings, our STS method consistently attains the highest success ratio and remains stable as speed and difficulty increase, whereas CTS and the Baseline show a clear drop in performance on faster and more complex terrains. This highlights both the advantage of the successive teacher–student schedule, which yields a stronger student policy than CTS, and the benefit of our adaptive gait-frequency action in the unified action space, which modulates the stepping rate according to commanded velocity and terrain type to maintain safe and precise foot placement.

\begin{table}[hb]
\centering
\caption{Ablation of loss terms for the reconstruction network (averaged over all terrain types).}
\scalebox{0.95}{
\begin{tabular}{lccc}
\toprule
\textbf{Component} 
  & \textbf{BCE} $\downarrow$
  & \textbf{DICE} ($\times 10^{-2}$) $\downarrow$
  & \textbf{MAE (cm)} $\downarrow$ \\
\midrule
\textbf{Our}        
  & \textbf{0.08$\pm$0.03} 
  & \textbf{1.81$\pm$0.31 } 
  & \textbf{2.64$\pm$0.12} \\
only-Dice           
  & 0.62$\pm$0.25 
  & 2.03$\pm$0.29 
  & 2.98$\pm$0.18 \\
only-BCE            
  & 0.18$\pm$0.08 
  & 2.15$\pm$0.21 
  & 2.91$\pm$0.18 \\
w/o edge branch     
  & 0.41$\pm$0.24 
  & 2.79$\pm$0.04 
  & 3.58$\pm$0.22 \\
\bottomrule
\end{tabular}
}
\label{tab:ablation_depth_preproc}
\end{table}

\subsection{Reconstruction Module Results}
Fig.~\ref{fig:pcd} shows reconstruction results of the perception module on three typical terrains: frontal stairs, a gap terrain, and side-facing stairs. 
In all cases, the network recovers a dense under-base height map that fills in regions heavily occluded in the raw height map by the legs. 
The reconstructed surfaces align well with the original geometry, preserving sharp edges, which provides a clean local structure for foothold selection.

Table~\ref{tab:ablation_depth_preproc} reports an ablation study of the loss terms used in the reconstruction network, averaged over all terrain types. 
The full model (\emph{Ours}), which uses an explicit edge branch supervised by both BCE and DICE losses, achieves the best scores in BCE, DICE, and MAE. 
Removing the edge branch (\emph{w/o edge branch}) leads to the largest errors, showing that explicitly modeling edges is crucial for capturing height discontinuities. 
Using only DICE (\emph{only-Dice}) degrades pixel-wise edge accuracy, while using only BCE (\emph{only-BCE}) harms region-level consistency under class imbalance. 
These results indicate that the edge branch and the combination of BCE and DICE losses play complementary roles in producing accurate and coherent height maps for downstream locomotion control.

The perception module produces height maps that capture both smooth regions and sharp discontinuities around gaps and stair edges. Integrated into the control pipeline, these height maps support precise foothold planning, enabling stable forward, backward, and turning motions in complex environments.

\subsection{Adaptive Gait Analysis}
We now examine the adaptive gait behavior enabled by the unified policy. Fig.~\ref{fig:gait_adaptation} illustrates an example trajectory where the robot moves across flat terrain, climbs stairs, turns, and descends stairs. The plot shows the commanded forward velocity alongside the corresponding gait frequency action over time.
On flat terrain, the gait frequency increases in response to the commanded forward velocity, resulting in smoother and more efficient walking. When the robot transitions to stairs or turns, the controller automatically adjusts the gait frequency to accommodate the new task. On stairs, the gait slows down to ensure precise and safe foot placement, while during turning, the frequency is modulated to maintain balance and coordination between the upper and lower body. Notably, the gait frequency can change abruptly to maintain stability.

Overall, the experiments indicate that the unified policy effectively fuses terrain-aware height map perception with proprioceptive feedback. By adjusting the gait frequency based on both high-level commands and local terrain conditions, the robot achieves flexible and robust locomotion across various scenarios. In particular, on challenging terrains such as stairs and gaps, the controller can adapt its gait rhythm to secure precise footholds, avoid missteps or collisions, and maintain stability throughout the motion.

\begin{figure}[hbt]
    \centering    \includegraphics[width=0.95\linewidth]{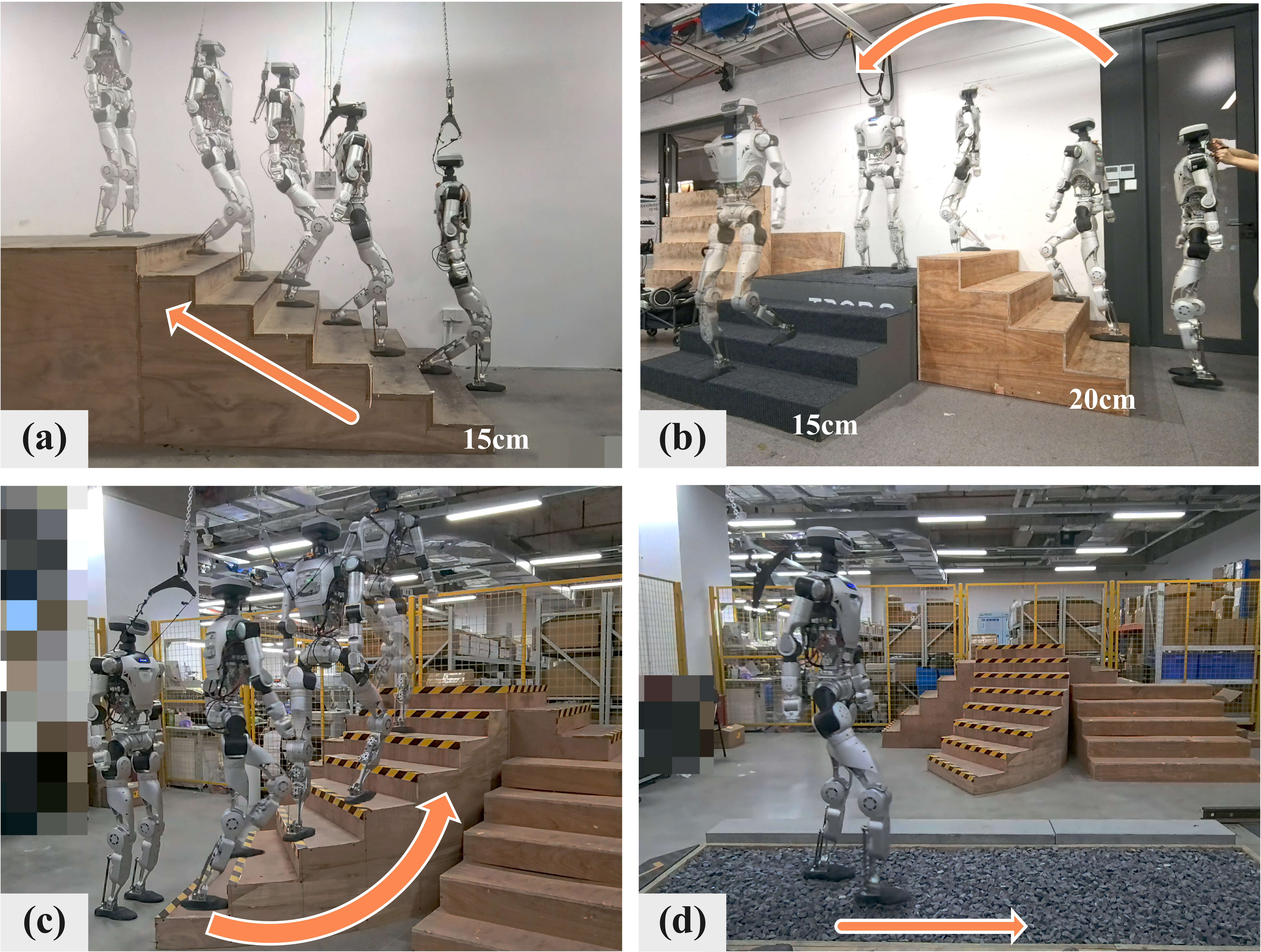}
    \setlength{\belowcaptionskip}{-15pt}  
    \caption{Real-world deployment on Oli: 
(a) climbing stairs backwards; 
(b) ascending and descending stairs forwards;
(c) climbing an unseen spiral staircase, demonstrating zero-shot generalization; 
and (d) traversing a patch of loose gravel.}
    \label{fig:robot_real_2}
\end{figure}

\subsection{Real-World Deployment}
We deploy the final policy on our full-sized humanoid robot Oli and evaluate it in a variety of real-world scenarios (Fig.~\ref{fig:robot_real_1} and Fig.~\ref{fig:robot_real_2}). 
The controller trained in simulation is run on hardware together with the on-board depth camera and under-base reconstruction module at 50\,Hz, without any task-specific retuning.

Outdoors, Oli climbs long flights of stairs both up and down while maintaining a stable, human-like gait. 
The robot climbs 15\,cm stairs forward and sideways, descends stairs backwards, and performs turning manoeuvres on a spiral staircase that is never seen during training.
This demonstrates zero-shot generalization of the perceptive locomotion and gait-adaptation framework to previously unseen stair geometries.
The policy also enables Oli to climb 20\,cm stairs, step over a 46\,cm gap and to traverse a patch of loose gravel, showing that the reconstructed under-base height maps and unified gait control can handle both sharp height discontinuities and irregular, deformable terrain. 
Across these diverse conditions, Oli remains balanced and tracks the commanded walking direction, demonstrating that the proposed perceptive locomotion framework transfers robustly from simulation to real hardware on complex stair and gap terrains.

\section{Conclusion}
We presented a perceptive humanoid locomotion framework that tightly couples under-base depth perception, unified gait and joint control, and a single-stage Successive Teacher--Student scheme. A single policy consumes local height maps and proprioception to jointly produce whole-body joint commands and gait frequency, enabling terrain-aware, gait-adaptive walking on complex stairs and gap terrains. Experiments in IsaacLab and on our full-sized humanoid, Oli, demonstrate stable omnidirectional locomotion, including forward, backward, and sideways stair traversal, gap crossing, and walking on loose outdoor surfaces.

In future work, we plan to extend this framework to higher-speed locomotion, more proactive obstacle avoidance, and autonomous waypoint tracking, aiming towards more versatile humanoid locomotion in diverse real-world environments.

\addtolength{\textheight}{-12cm}   









\bibliographystyle{IEEEtran} 
\bibliography{IEEEabrv,IEEEexample}

@article{legged-robots,
  title={Learning agile and dynamic motor skills for legged robots},
  author={Hwangbo, Jemin and Lee, Joonho and Dosovitskiy, Alexey and Bellicoso, Dario and Tsounis, Vassilios and Koltun, Vladlen and Hutter, Marco},
  journal={Science Robotics},
  volume={4},
  number={26},
  pages={eaau5872},
  year={2019},
  publisher={American Association for the Advancement of Science}
}

@article{li2025reinforcement,
  title={Reinforcement learning for versatile, dynamic, and robust bipedal locomotion control},
  author={Li, Zhongyu and Peng, Xue Bin and Abbeel, Pieter and Levine, Sergey and Berseth, Glen and Sreenath, Koushil},
  journal={The International Journal of Robotics Research},
  volume={44},
  number={5},
  pages={840--888},
  year={2025},
  publisher={SAGE Publications Sage UK: London, England}
}

@article{radosavovic2024real,
  title={Real-world humanoid locomotion with reinforcement learning},
  author={Radosavovic, Ilija and Xiao, Tete and Zhang, Bike and Darrell, Trevor and Malik, Jitendra and Sreenath, Koushil},
  journal={Science Robotics},
  volume={9},
  number={89},
  pages={eadi9579},
  year={2024},
  publisher={American Association for the Advancement of Science}
}

@article{gu2024humanoid,
  title={Humanoid-gym: Reinforcement learning for humanoid robot with zero-shot sim2real transfer},
  author={Gu, Xinyang and Wang, Yen-Jen and Chen, Jianyu},
  journal={arXiv preprint arXiv:2404.05695},
  year={2024}
}

@article{xie2025humanoid,
  title={Humanoid whole-body locomotion on narrow terrain via dynamic balance and reinforcement learning},
  author={Xie, Weiji and Bai, Chenjia and Shi, Jiyuan and Yang, Junkai and Ge, Yunfei and Zhang, Weinan and Li, Xuelong},
  journal={arXiv preprint arXiv:2502.17219},
  year={2025}
}

@inproceedings{zhuang2024humanoid,
  title={Humanoid Parkour Learning},
  author={Zhuang, Ziwen and Yao, Shenzhe and Zhao, Hang},
  booktitle={Conference on Robot Learning ({CoRL})},
  year={2024}
}

@article{sun2025dpl,
  title={DPL: Depth-only Perceptive Humanoid Locomotion via Realistic Depth Synthesis and Cross-Attention Terrain Reconstruction},
  author={Sun, Jingkai and Han, Gang and Sun, Pihai and Zhao, Wen and Cao, Jiahang and Wang, Jiaxu and Guo, Yijie and Zhang, Qiang},
  journal={arXiv preprint arXiv:2510.07152},
  year={2025}
}

@article{long2024learninghumanoid,
  title={Learning Humanoid Locomotion with Perceptive Internal Model},
  author={Long, Junfeng and Ren, Junli and Shi, Moji and Wang, Zirui and Huang, Tao and Luo, Ping and Pang, Jiangmiao},
  journal={IEEE International Conference on Robotics and Automation ({ICRA})},
  year={2025}
}

@article{luo2024pie,
  title={Pie: Parkour with implicit-explicit learning framework for legged robots},
  author={Luo, Shixin and Li, Songbo and Yu, Ruiqi and Wang, Zhicheng and Wu, Jun and Zhu, Qiuguo},
  journal={IEEE Robotics and Automation Letters},
  year={2024},
  publisher={IEEE}
}

@inproceedings{wang2025beamdojo,
  title     = {BeamDojo: Learning Agile Humanoid Locomotion on Sparse Footholds},
  author    = {Wang, Huayi and Wang, Zirui and Ren, Junli and Ben, Qingwei and Huang, Tao and Zhang, Weinan and Pang, Jiangmiao},
  booktitle = {Robotics: Science and Systems ({RSS})},
  year      = {2025},
}

@article{gu2024advancing,
  title={Advancing humanoid locomotion: Mastering challenging terrains with denoising world model learning},
  author={Gu, Xinyang and Wang, Yen-Jen and Zhu, Xiang and Shi, Chengming and Guo, Yanjiang and Liu, Yichen and Chen, Jianyu},
  journal={arXiv preprint arXiv:2408.14472},
  year={2024}
}

@inproceedings{duan2024learning,
  title={Learning vision-based bipedal locomotion for challenging terrain},
  author={Duan, Helei and Pandit, Bikram and Gadde, Mohitvishnu S and Van Marum, Bart and Dao, Jeremy and Kim, Chanho and Fern, Alan},
  booktitle={2024 IEEE International Conference on Robotics and Automation (ICRA)},
  pages={56--62},
  year={2024},
  organization={IEEE}
}

@article{wang2024cts,
  title={Cts: Concurrent teacher-student reinforcement learning for legged locomotion},
  author={Wang, Hongxi and Luo, Haoxiang and Zhang, Wei and Chen, Hua},
  journal={IEEE Robotics and Automation Letters},
  year={2024},
  publisher={IEEE}
}

@inproceedings{xue2025unified,
  title={A Unified and General Humanoid Whole-Body Controller for Fine-Grained Locomotion}, 
  author={Xue, Yufei and Dong, Wentao and Liu, Minghuan and Zhang, Weinan and Pang, Jiangmiao},
  booktitle={Robotics: Science and Systems (RSS)},
  year={2025},
  }

@article{lu2025contrastive,
  title={Contrastive Representation Learning for Robust Sim-to-Real Transfer of Adaptive Humanoid Locomotion},
  author={Lu, Yidan and Yang, Rurui and Kou, Qiran and Chen, Mengting and Fan, Tao and Cui, Peter and Dong, Yinzhao and Lu, Peng},
  journal={arXiv preprint arXiv:2509.12858},
  year={2025}
}

@inproceedings{siekmann2021blind,
  title={Blind Bipedal Stair Traversal via Sim-to-Real Reinforcement Learning},
  author={Siekmann, Jonah and Green, Kevin and Warila, John and Fern, Alan and Hurst, Jonathan},
  booktitle={Robotics: Science and Systems},
  year={2021}
}

@article{wang2025more,
  title={MoRE: Mixture of Residual Experts for Humanoid Lifelike Gaits Learning on Complex Terrains},
  author={Wang, Dewei and Wang, Xinmiao and Liu, Xinzhe and Shi, Jiyuan and Zhao, Yingnan and Bai, Chenjia and Li, Xuelong},
  journal={arXiv preprint arXiv:2506.08840},
  year={2025}
}

@article{schulman2017proximal,
  title={Proximal policy optimization algorithms},
  author={Schulman, John and Wolski, Filip and Dhariwal, Prafulla and Radford, Alec and Klimov, Oleg},
  journal={arXiv preprint arXiv:1707.06347},
  year={2017}
}

@misc{oli,
  title        = {\textcolor{black}{Limx Oli}},
  year         = {\textcolor{black}{2025}},
  howpublished = {\url{https://www.limxdynamics.com/en/oli}},
  note         = {\textcolor{black}{Accessed: July 30, 2025}}
}

@article{mittal2025isaac,
  title={Isaac Lab: A GPU-Accelerated Simulation Framework for Multi-Modal Robot Learning},
  author={Mittal, Mayank and Roth, Pascal and Tigue, James and Richard, Antoine and Zhang, Octi and Du, Peter and Serrano-Mu{\~n}oz, Antonio and Yao, Xinjie and Zurbr{\"u}gg, Ren{\'e} and Rudin, Nikita and others},
  journal={arXiv preprint arXiv:2511.04831},
  year={2025}
}

@inproceedings{li2020dice,
  title={Dice loss for data-imbalanced NLP tasks},
  author={Li, Xiaoya and Sun, Xiaofei and Meng, Yuxian and Liang, Junjun and Wu, Fei and Li, Jiwei},
  booktitle={Proceedings of the 58th annual meeting of the association for computational linguistics},
  pages={465--476},
  year={2020}
}

@article{tong2024advancements,
  title={Advancements in humanoid robots: A comprehensive review and future prospects},
  author={Tong, Yuchuang and Liu, Haotian and Zhang, Zhengtao},
  journal={IEEE/CAA Journal of Automatica Sinica},
  volume={11},
  number={2},
  pages={301--328},
  year={2024},
  publisher={IEEE}
}

@article{yang2022real,
  title={Real-time neural dense elevation mapping for urban terrain with uncertainty estimations},
  author={Yang, Bowen and Zhang, Qingwen and Geng, Ruoyu and Wang, Lujia and Liu, Ming},
  journal={IEEE Robotics and Automation Letters},
  volume={8},
  number={2},
  pages={696--703},
  year={2022},
  publisher={IEEE}
}

@article{hoeller2022neural,
  title={Neural scene representation for locomotion on structured terrain},
  author={Hoeller, David and Rudin, Nikita and Choy, Christopher and Anandkumar, Animashree and Hutter, Marco},
  journal={IEEE Robotics and Automation Letters},
  volume={7},
  number={4},
  pages={8667--8674},
  year={2022},
  publisher={IEEE}
}

\end{document}